\documentclass[times,twocolumn,final,authoryear]{elsarticle}

\usepackage{ycviu}
\usepackage{framed,multirow}

\usepackage{amssymb}
\usepackage{latexsym}

\usepackage{url}
\usepackage{xcolor}
\usepackage{amsmath}
\usepackage{booktabs} 
\usepackage{tabularx}
\usepackage{hyperref}
\definecolor{newcolor}{rgb}{.8,.349,.1}
\usepackage{appendix}

\journal{Computer Vision and Image Understanding}

\begin{document}

\clearpage
\thispagestyle{empty}
\ifpreprint
  \vspace*{-1pc}
\fi

\begin{frontmatter}

\title{Vessel Re-identification and Activity Detection in Thermal Domain for Maritime Surveillance}

\author[1]{Yasod \snm{Ginige}} 
\author[1]{Ransika \snm{Gunasekara}}
\author[1]{Darsha \snm{Hewavitharana}}
\author[1]{Manjula \snm{Ariyarathne}}
\author[1]{Ranga \snm{Rodrigo}}
\author[1]{Peshala \snm{Jayasekara}\corref{cor1}}
\cortext[cor1]{Corresponding author: 
  Tel.: +94-711273844; }
\ead{peshala@uom.lk}

\address[1]{University of Moratuwa, Katubedda, Colombo (10400), Sri Lanka}

\received{1 May 2013}
\finalform{10 May 2013}
\accepted{13 May 2013}
\availableonline{15 May 2013}
\communicated{S. Sarkar}

\begin{abstract}
Maritime surveillance is vital to mitigate illegal activities such as drug smuggling, illegal fishing, and human trafficking. Vision-based maritime surveillance is challenging mainly due to visibility issues at night which results in failures in re-identifying vessels and detecting suspicious activities. In this paper, we introduce a thermal, vision-based approach for maritime surveillance with object tracking, vessel re-identification, and suspicious activity detection capabilities. For vessel re-identification, we propose a novel viewpoint-independent algorithm which compares features of the sides of the vessel separately (separate side-spaces) leveraging shape information in the absence of color features. We propose techniques to adapt tracking and activity detection algorithms for the thermal domain and train them using a thermal dataset we created. This dataset will be the first publicly available benchmark dataset for thermal maritime surveillance. Our system is capable of re-identifying vessels with an 81.8\% Top1 score and identifying suspicious activities with a 72.4\% frame mAP score; a new benchmark for each task in the thermal domain.

\end{abstract}


\end{frontmatter}


\section{Introduction}\label{sec1}
Illicit maritime activities such as smuggling, illegal fishing, and human trafficking are major threats, especially during the night.
Maritime surveillance, a preventive measure and a deterrent in the face of this threat, involves detection, tracking, re-identification, and blacklisting vessels. 
Vision based maritime surveillance is challenging due to visibility issues at night, water body reflections, and extreme weather conditions.

\begin{figure*}[t]
    \centering
    \includegraphics[width=\linewidth]{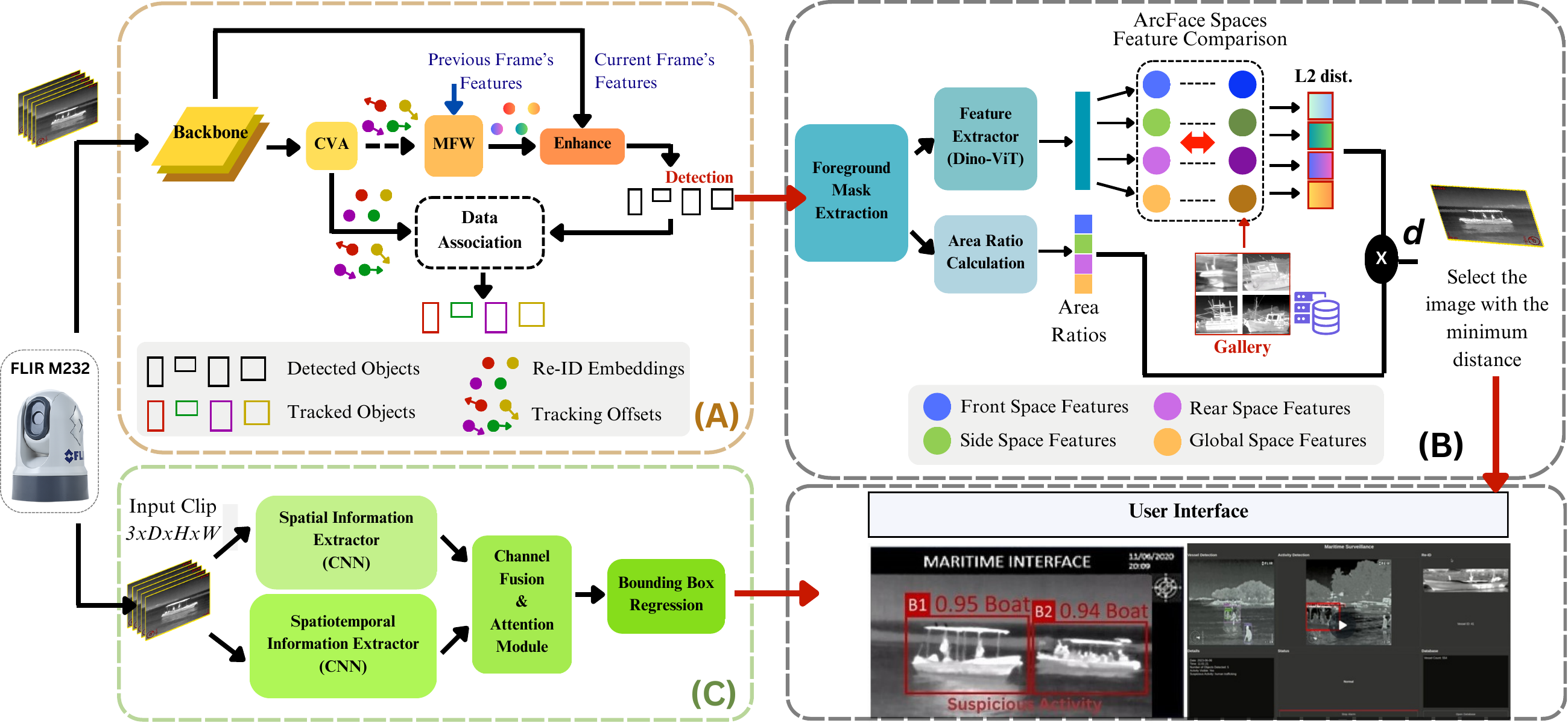}
    \caption{Overall structure of the proposed surveillance system. Object detection and tracking algorithms (A) identify maritime objects in each frame, and crop and feed them to the re-identification algorithm (B). It extracts features from each visible side of the vessel and compares them with the dynamic database. Activity detection algorithm (C) detects suspicious activities. Sections 3.1, 3.2, and 3.3 further discuss (A), (B), and (C) subsystems, respectively.}
    \label{big_picture}
\end{figure*}

In non-maritime RGB domains, however, video-based surveillance is well established, especially in building and road traffic surveillance systems. Human tracking, counting, identification, authorization, and hazard detection are common use cases of surveillance systems~[\cite{vid_sr_1,vid_sr_2,vid_sr_3}]. In road traffic surveillance, cameras detect speeding and other rule violations, monitor traffic conditions, and provide parking assistance~[\cite{vid_sr_0,vid_sr_4}]. In face recognition and person re-identification, existing methods lock on to facial landmarks, body structure, and clothing~[\cite{ma2012bicov,cheng2020face}]. Many methods in face recognition, person re-identification, and traffic detection domains use fine-grained RGB structures such as specific features of the human face, patterns and color in clothing, and number plates of vehicles. These methods that work in the RGB domain fail in gray-scale maritime thermal videos, particularly, as thermal images capture a different set of features and due to the absence of distinctive features as mentioned above.


In maritime environments, vessels can appear in entirely different views, unlike in face re-identification. The features of each side of these vessels are vastly different from the other sides. Therefore, the algorithms should be able to build a feature vector paying attention to vessel orientation and comparing related features with other vessels. Furthermore, checking which sides are visible in a vessel image is itself a challenging problem that should be solved prior to the orientation-wise feature extraction. 

In this paper, we aim at creating a maritime surveillance system mainly focusing on robust vessel re-identification that benefits from the nature of features visible in maritime thermal images. Our system (Fig.~\ref{big_picture}) comprises three main subsystems, namely, maritime vessel tracking, vessel re-identification, and maritime activity detection. In the tracking subsystem, we train an algorithm to track maritime objects in the
thermal domain to withstand night and extreme weather conditions. The re-identification subsystem builds a view-weighted feature vector that captures all visible sides of the vessel using an encoder-decoder foreground segmenter and a ViT-based part (view) attention network similar to SPAN~[\cite{SPAN}]. This feature vector,  using an ArcFace loss~[\cite{deng2019arcface}], identifies vessels using a dynamic database, irrespective of the orientation (achieving viewpoint independence) of the query image by focusing on distinct shapes on each visual side of the vessel. This compensates for the absence of color and fine features in the thermal domain. The maritime activity detection subsystem incorporates both spatial and temporal action localization trained on maritime thermal images. We adopt the YOWO~[\cite{kopuklu2019you}] algorithm in this work, which has been exclusively used in RGB domain, to detect two representative maritime activities.

Our main contributions are as follows:
\begin{itemize}
\item {\bf Viewpoint-independent novel re-identification algorithm:} The algorithm focuses on the shape of vessels in the absence of color and intricate features in the thermal domain and compares feature vectors in each side-space, separately. The algorithm outperforms the mAP score of the  SPAN model~[\cite{SPAN}] by 32\% in the thermal domain. 

\item {\bf Creation of a thermal maritime dataset:} The annotated dataset contains video footage of maritime vessels, jet-skies, and human activities with COCO annotations~[\cite{coco}]. It also contains images of 40 small vessels and 32 large vessels from different viewpoints. The dataset can be used for detection and tracking, re-identification, and activity detection tasks. To the best of our knowledge, this is the first public dataset created for maritime surveillance in the thermal domain.\\ Link: \href{https://hevidra.github.io/}{https://hevidra.github.io/}

\item {\bf Detection and tracking in the thermal domain:} We adapted the TraDes~[\cite{wu2021track}] algorithm, which is originally trained on RGB data, to track maritime objects such as vessels, ships, jet-skies, and humans in the thermal domain. The algorithm was fine-tuned using the COCO and the Singapore Maritime Dataset (SMD)~[\cite{b15}]. We tuned detection and tracking thresholds and achieved a 61.2\% MOTA score.

\item {\bf Activity detection in the thermal domain:} We adapted the YOWO~[\cite{kopuklu2019you}] algorithm, which was originally trained in the RGB domain, to detect activities such as possible human trafficking and swimming in the thermal domain. We re-trained the algorithm on our dataset and tuned hyper-parameters to obtain a frame mAP score of 62.45\%, demonstrating promising detection of target activities.
\end{itemize}

\section{Related Work}\label{sec2}
In this section, we explore literature under three main areas: vessel re-identification, object tracking and activity detection.\\

\subsection{Re-identification}
 
Re-identification is the process of identifying the same object or individual across different scenes, which typically involves matching features across images or video frames captured at different times and locations. It extends into different subdomains including face, human, objects, and vehicle re-identification. Face re-identification methods pay attention to specific features such as the iris, dimension of the face, nose, lips, and the color of the eyeball~[\cite{cheng2020face}]. Due to genetic organizations, each human has a unique combination of these features which makes the face re-identification possible. 
However, it doesn't facilitate re-identification from different angles of the face as the algorithm expects the full frontal view of the face. 
In human re-identification tasks, algorithms pay attention to the whole body in addition to the face. Thus, there are other features such as height, shape, body language, and colors of the clothes taken into consideration~[\cite{ma2012bicov}]. More recent algorithms are capable of re-identifying despite different orientations~[\cite{shi2022iranet,bansal2022cloth}]. However, in the thermal domain, human re-identification has been a challenging task; some approaches get the guidance from a visible model to train the thermal model~[\cite{ye2018visible}], while fully thermal approaches suffer from low accuracy~[\cite{humanIR}].

In vehicle re-identification, the main challenge is the variability in viewpoint and the high similarity among vehicles of the same category. Recent work has introduced several techniques that address the viewpoint variation by considering the camera's perspective~[\cite{chu2019vehicle, qiao2020marine, zhou2018aware}]. These methods aim to learn the similarities and differences between images captured from different viewpoints by using triplet loss across extracted feature mappings. It enables accurate re-identification across various camera angles. However, these methods have the advantage of color features markedly absent in thermal domain. Furthermore, they have not been tested under poor visibility conditions, such as night-time and bad weather. Thermal domain vehicle re-identification is not well explored. Eleni \emph{et al.}~[\cite{kamenou2022closing}] have proposed a cross-domain model and tried to learn sharable features in both visible and IR domains. The model contains a shareable network followed by two separate streams for two domains increasing the computational complexity. The same authors propose a domain generalization approach for multi-modal vehicle re-identification based on meta-learning~[\cite{kamenou2023meta}] using RGB, near-IR, and IR domains. However, both methods share visible domain features when training, possibly paying less attention to shapes than color features. Furthermore, they expect the model to see images of query vehicles from a similar orientation in the inference, i.e., the method does not force the model to learn orientation based feature extraction and identity classification. Nevertheless, for a maritime surveillance system, we cannot guarantee that the gallery contains images of a vessel from all the orientations (front, side, rear, front-and-side,...). Hence, the re-identification algorithm should be robust for appearances from different orientations. Chen \emph{et al.} have proposed a viewpoint aware re-identification algorithm, SPAN~[\cite{SPAN}] for RGB domain, which has the additional advantage of color features compared to the thermal domain. It pays less attention to the unique shapes of the vehicle (due to minor modifications) when comparing two vehicles of the same model. 
In our case, the domain differs from RGB to thermal, and minor changes in the shape of vessels are significant when identifying vessels.

To the best of our knowledge, there is no work done in the thermal domain for maritime vessel re-identification. In this paper, we combine thermal images with orientation based feature extraction and identity classification, solving both visibility issues and the orientation issues.

\subsection{Object Tracking}
Object tracking is the automated process of locating and following objects of interest in images or videos. Conventional approaches such as~[\cite{kim2015multiple, tang2017multiple,schulter2017deep}] use two stages for detection and tracking, consuming more computational power and time. In these algorithms, a backbone model is used to detect objects, and then, a separate association algorithm builds tracklets between adjacent frames using those detections. Therefore, they cannot usually be used for real time object tracking due to heavy processing.

To overcome these challenges, recent work has moved towards joint detection and tracking approaches, where we detect and track using a single backbone model. Other than real time processing, we need to keep a clean tracklet for each detected object throughout the frames as we feed these objects to the re-identification algorithm at predefined time steps. If the tacking algorithm cannot maintain a consistent identification (ID) for an object, the re-identification algorithm will be triggered upon every new object ID, causing redundant computations. Simple Online Realtime Tracker (SORT)~[\cite{bewley2016simple}] uses a Kalman filter to estimate the object's location from the previous frame and leverages measurements with uncertainty to estimate the current states. 
Deep SORT~[\cite{wojke2017simple}], an extension of SORT, compares the appearance of new detections with previously tracked objects within each track to assist data association using a re-identification based approach. 
However, in these methods, detection is independently predicted without tracking assistance that prevents a possible accuracy increment. This leads to frequent possible ID updates of detected objects in occluded or unclear scenarios. TraDeS, introduced by Jialian Wu \emph{et al.}~[\cite{wu2021track}], presents an online multi-object tracking algorithm that integrates object detection and tracking to achieve robust and accurate tracking performance using CenterNet~[\cite{duan2019centernet}] as the backbone. It uses a peer supporting technique where features extracted from detection helps the tracking objective, while tracking offsets predicted in the detection stage and feature of previous frames enhance features of the current frame to help the detection objective. Although it can be used for real-time, accurate tracking tasks, it has not been tested in the thermal domain. In this paper, we train and evaluate the TraDeS algorithm on thermal data and provide a comprehensive results analysis to prove the validity of the algorithm in thermal domain object tracking.

\subsection{Activity Detection and Localization}

Vision based activity detection uses cameras to capture a video feed and processes it sequentially to identify activities ~[\cite{oneata2014efficient,weinzaepfel2015learning}]. Recent work has paid attention towards two stream localization which combines both spatial and temporal streams, thereby improving the detection and classification of actions within a video~[\cite{peng2016multi}]. However, only a few methods provide both online and realtime activity detection while maintaining a higher accuracy. Singh \emph{et al.}~[\cite{singh2017online}] focus on online real-time action localization and prediction in real-time. This method has the capability to localize actions and predict upcoming actions, demonstrating the potential of predictive modeling. However, YOWO~[\cite{kopuklu2019you}], proposed by Okan \emph{et al.}, is a comparatively low weight method with both online and real time processing capabilities and higher accuracy.  It uses a unified CNN architecture for real-time spatiotemporal action localization using only a single pass through the network. This allows to process the video with a higher frames-per-second (fps) which is a considerable improvement over previous methods that require multiple iterations or separate processes for different tasks. Nevertheless, YOWO is utilized only in the RGB domain and is not tested for detecting suspicious activities such as possible human trafficking. In our work, we show that YOWO can be adapted for the thermal domain by retraining and adjusting hyper-parameters and sets a new benchmark for thermal activity detection.

\section{Methodology}\label{sec3}

The framework proposed in this study comprises three primary subsystems: object tracking, vessel re-identification, and activity detection, as depicted in Fig.~\ref{big_picture}. The thermal video feed captured by the camera is directed towards the object tracking and activity detection subsystems. Subsequently, the tracking subsystem outputs identified objects, which are then forwarded to the re-identification subsystem. The outputs generated by all three subsystems are integrated into a user interface, facilitating the visualization of detected marine vessels, associated activities, and the corresponding re-identification results. The following sections explain each subsystem in detail.

\subsection{Object Tracking for Bounding Box Extraction}

\begin{figure}[htbp]
    \centering
    \includegraphics[width=\linewidth]{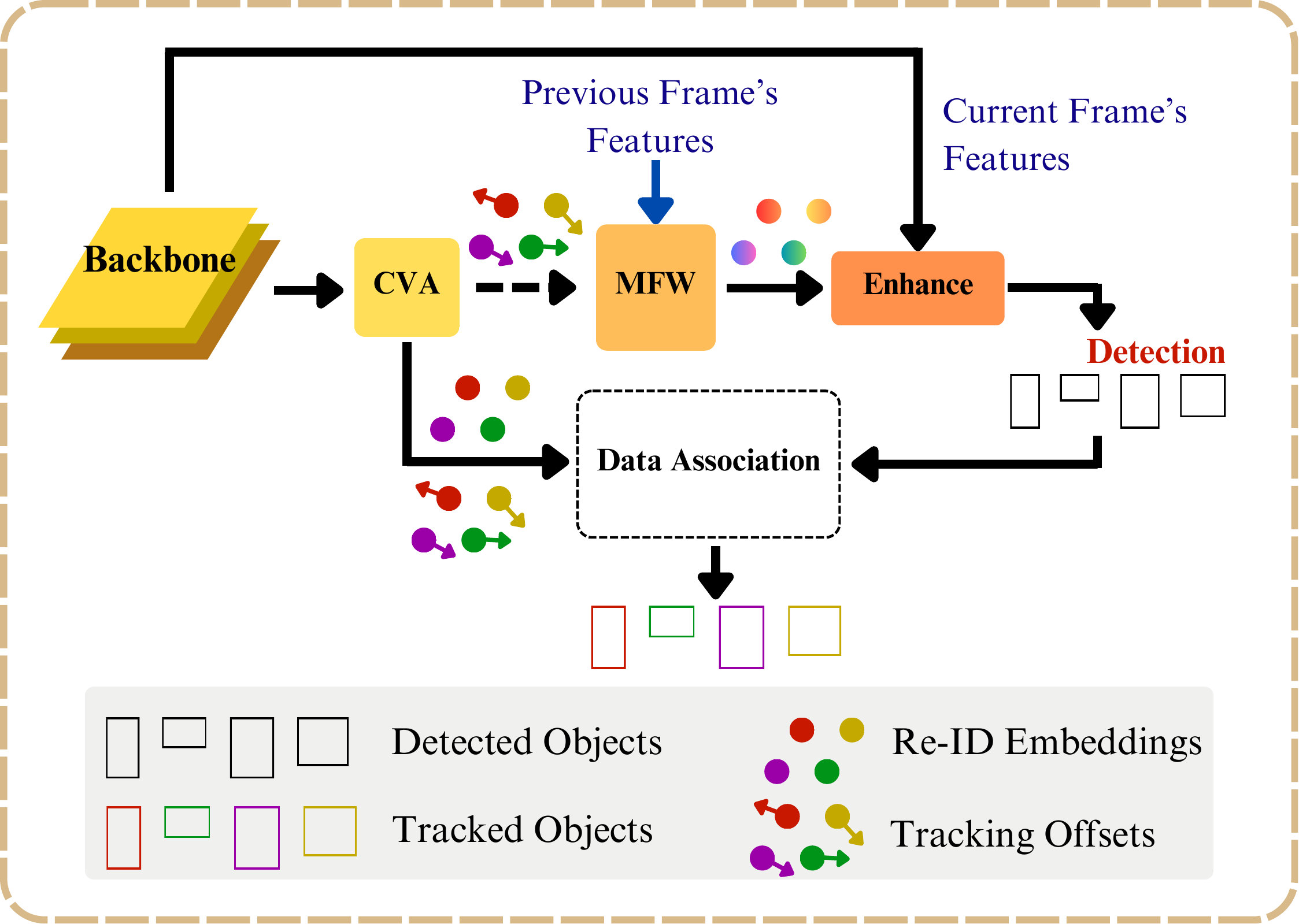}
    \vspace{9pt}
    \caption{Object tracking model architecture}
    \label{track_picture}
\end{figure}

For object tracking, we adapted TraDeS~[\cite{wu2021track}] algorithm for the thermal domain. While many existing detection and tracking approaches conduct independent detection without incorporating tracking input, this method integrates tracking cues into the detection process to enhance performance in challenging scenarios, and thereby improving tracking outcomes. First, we use the DLA-34 model~[\cite{yu2018deep}] as the backbone for the feature extraction of input frames. Next, we use two modules to optimize object detection and tracking using the outcomes of each other (Fig.~\ref{track_picture}). The Cost Volume based Association (CVA) module is used to generate embeddings and derive object motions to improve object tracking accuracy. Then, we use the Motion-guided Feature Warper (MFW) module to enhance the object features in the next frame based on the CVA outcomes. More specifically, MFW enhances the feature vector of the current frame based on the tracking history of past frames. This improves the performance of the algorithms, especially when the current frame is occluded. It utilizes tracking cues obtained from the CVA, and propagates them to enhance object features to improve the detection accuracy. Next, object detection is done using CenterNet~[\cite{duan2019centernet}] in the current frame and is associated using a two-round data association technique. In the first round, objects are mapped to the closest tracklet. If it fails, cosine similarity between unmatched tracklet embeddings and the object feature embedding is considered.   This method was chosen because of the model’s ability to integrate detecting, segmenting, and tracking in a single network, which reduces the processing time and improves overall accuracy and efficiency. 
In addition, the model has better real-time tracking of multiple objects compared to previous models discussed in section~\ref{sec2}, and it is robust to occlusions and appearance changes.

We adapt the TraDeS algorithm for the thermal domain by retraining it using thermal data. We replicate the thermal channel into 3 channels (RGB) when feeding it to the algorithm. The training process is further discussed in section~\ref{sec4.2}.

\begin{figure*}[t]
    \centering
    \includegraphics[width=0.8\linewidth]{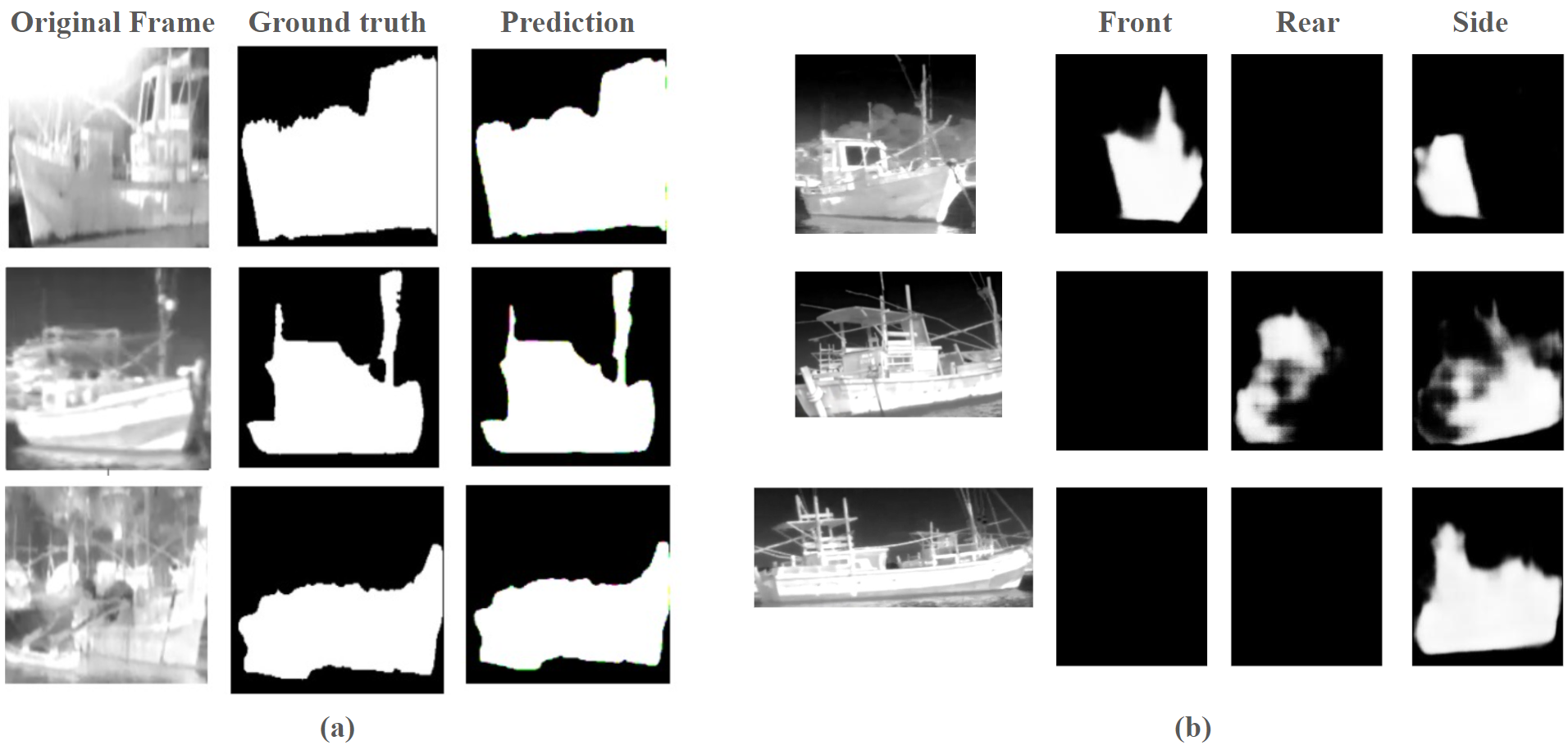}
    \caption{(a) Results of the encoder-decoder model used in foreground extracting. (b) Masks generated by SPAN for front, side and rear views using the foreground mask.}
    \label{fig_enc_dec}
\end{figure*}
\subsection{Vessel Re-identification} \label{subsec3.2}
In vessel re-identification, our target is to extract the identity of a given query image using a set of gallery images (the database). 
Here, the main challenges are the lack of available data in thermal domain for marine vessels and the change of features for each vessel with different camera viewpoint. 
To tackle these issues, we used a model robust to the viewpoint, which can generalize well with a small amount of data. As the first step, we mask out the foreground (the vessel) from a given frame. Since thermal images do not contain color features and the intensity distribution of the foreground and the background are similar, conventional algorithms such as GrabCut~[\cite{rother2004grabcut}] do not perform well in this task. As a solution, we used an encoder-decoder architecture (Appendix: Table~\ref{tab:enc_dec}) with residual connections to build the foreground mask of a given frame. 
We annotated foreground masks of 300 images as ground truth labels, and trained the model using those frames as input. 
Then, as shown in Fig.~\ref{fig_enc_dec}, the trained model was tested on previously unseen data, demonstrating its capability to accurately mask out foreground elements with complex viewpoints and intensity variations. Therefore, we propose this encoder-decoder architecture as a foreground extractor, specifically for colorless images, given that it can be trained on relevant data.

\begin{figure*}[t]
    \centering
    \includegraphics[width=0.8\linewidth]{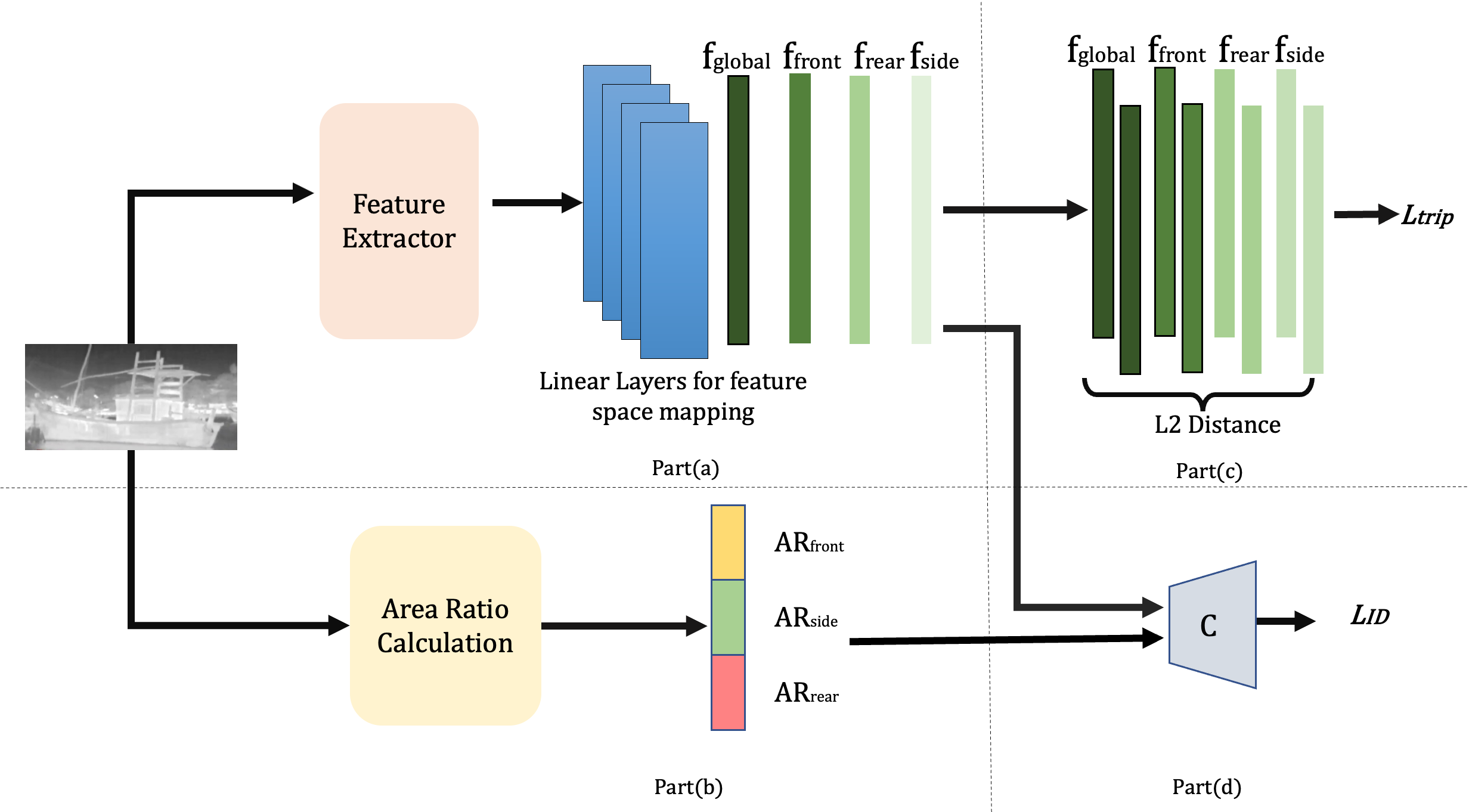}
    \caption{Re-identification subsystem. (a) extracts features using the vision transformer and maps them to four latent spaces. (b) calculates the area ratios of each visual side of the vessel. (c) and (d) calculate triplet loss and the ID loss, respectively.}
    \label{fig_reid_arch}
\end{figure*}

\begin{figure*}[t]
    \centering
    \includegraphics[width=0.9\linewidth]{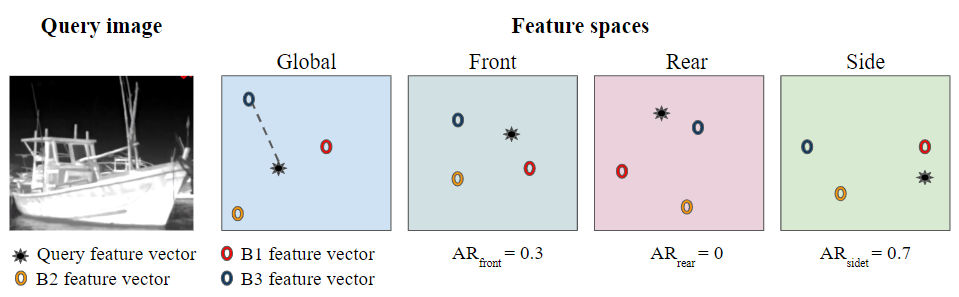}
    \caption{L2 distance calculation in separate spaces. According to the given vessel, more attention should be paid to distances in the side space while completely neglecting the rear space. Since B2 is the closest in the side space and comparatively close in the front space, the best match will be B2 for the given query image. }
    \label{fig_feature_spaces}
\end{figure*}

Next, the extracted foreground is fed to the identification model, where we use an architecture as shown in Fig.~\ref{fig_reid_arch}.(a) that extracts the features of a given vessel. 
As the extractor, we used the pre-trained Dino ViT transformer model presented by Mathilde Caron \emph{et al.}~[\cite{caron2021emerging}]. As shown in the Fig.~\ref{fig_reid_arch}.(c), we use four parallel linear layers to map the extracted feature vector to four latent spaces, namely global, front, rear, and side. 
We use these latent spaces to train the model to recognize vessel identities in different viewpoints since the vessel's features drastically change with the viewpoint. 
To get a better feature distribution, an ArcFace~[\cite{deng2019arcface}] mapping is used in each space. ArcFace is a feature analyzing technique that maps feature vectors onto a hypersphere, enhancing discrimination between different identities by maximizing inter-class variance while minimizing intra-class variance. It achieves state-of-the-art performance in face recognition tasks by embedding faces into a compact feature space. Next, we calculate L2 distances to each vessel in the database, in each space. These distances are multiplied by area ratios to embed the viewpoint information to the result and suppress erroneous information given from spaces corresponding to self-occluded views (Fig.~\ref{fig_feature_spaces}) as given in eq.~(\ref{eq:1}). Finally, we sort the total distances in ascending order and select the first identity as the match for the query image.

\begin{equation}
\label{eq:1}
\begin{aligned}
& \mathrm{Distance_{total}}(\mathrm{ID,Image}) = \\
& \quad \left\{ \mathrm{Distance_{global}(ID,Image)} \right. \\
& \quad + \mathrm{Distance_{front}}(\mathrm{ID,Image}) \cdot \mathrm{AR_{front}} \\
& \quad + \mathrm{Distance_{side}(ID,Image)} \cdot \mathrm{AR_{side}} \\
& \quad + \mathrm{Distance_{rear}(ID,Image)} \cdot \left.\mathrm{AR_{rear}}\right\}/2
\end{aligned}
\end{equation}

\begin{figure}[h!]
    \centering
    \includegraphics[width=\linewidth]{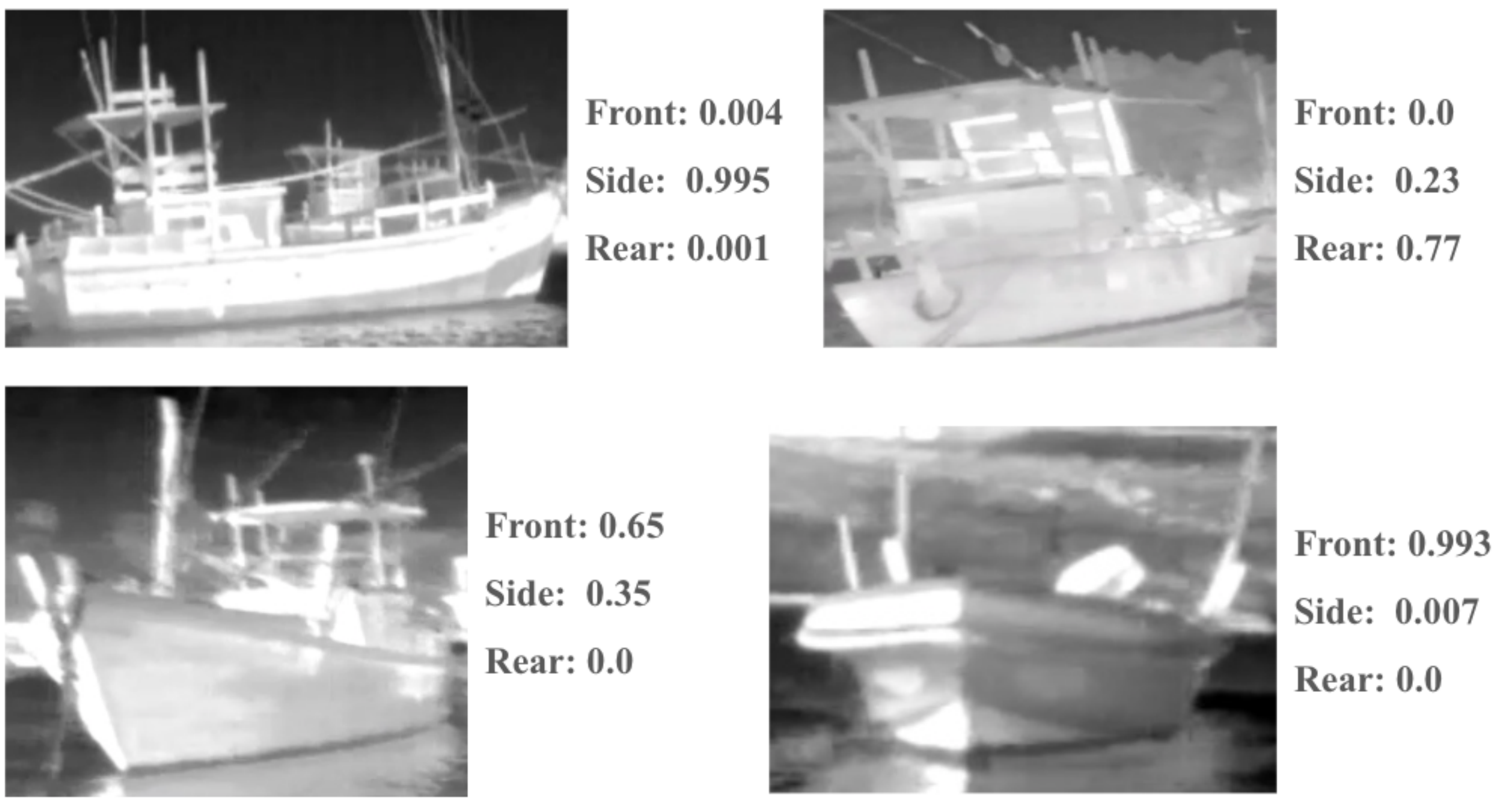}
    \vspace{4pt}
    \caption{Mapping visual orientations to numbers: calculating the area ratio vector of the Fig.~\ref{fig_reid_arch}.(b). The algorithm accurately identifies visible sides and assigns accurate values considering the area ratios.}
    \label{area_visual}
\end{figure}

\hfill \break
Inspired by the SPAN model, we calculate area ratios to generate masks of different viewpoints as shown in Fig.~\ref{fig_enc_dec}.(b). 
The ratio of each side is calculated using eq.~(\ref{eq:area_ratio}) and the qualitative evidence are shown in Fig.~\ref{area_visual}. 

\begin{equation}
\label{eq:area_ratio}
\begin{aligned}
\mathrm{AR_{side X}} &= \frac {\text{Area of the sideX view mask}}{\text{Area of the foreground mask}}, \\
& \text{where sideX} \in \{\text{front}, \text{side}, \text{rear}\}.
\end{aligned}
\end{equation}

When training the re-identification model, we freeze the area ratio calculation and fine-tune the linear layers to map the extracted features to the four spaces. We use identity classification and triplet loss in the training process. 

\subsubsection{Identity Classification Loss}
In the identity classification, after the features are mapped to the four spaces, we use ArcFace mapping to get the cosine distance of each viewpoint space to calculate the confidence of each space and calculate the confidence as given in eq.~(\ref{eq:1}). Then we use the cross-entropy loss as the identity classification loss.

\subsubsection{Triplet Loss}
 To promote discrimination and effective feature learning, we use triplet loss with Euclidean distance on the features mapped to the four spaces after the primary feature extraction. For the negative and positive samples, we use sample thermal images for each vessel identity manually, to make sure the model is robust to different viewpoints.

The total loss is calculated as in eq.~(\ref{eq:2}),
\begin{equation}
\label{eq:2}
\begin{aligned}
L_{\mathrm{Total}} = \lambda_{\mathrm{ID}} \cdot L_{\mathrm{ID}} +  \lambda_{\mathrm{Triplet}} \cdot L_{\mathrm{Triplet}}
\end{aligned}
\end{equation}
where  $\lambda_{\mathrm{ID}}$ and  $\lambda_{\mathrm{Triplet}}$ are hyper parameters.

\subsection{Activity Detection}

\begin{figure*}[htbp]
    \centering
    \includegraphics[width=0.8\linewidth]{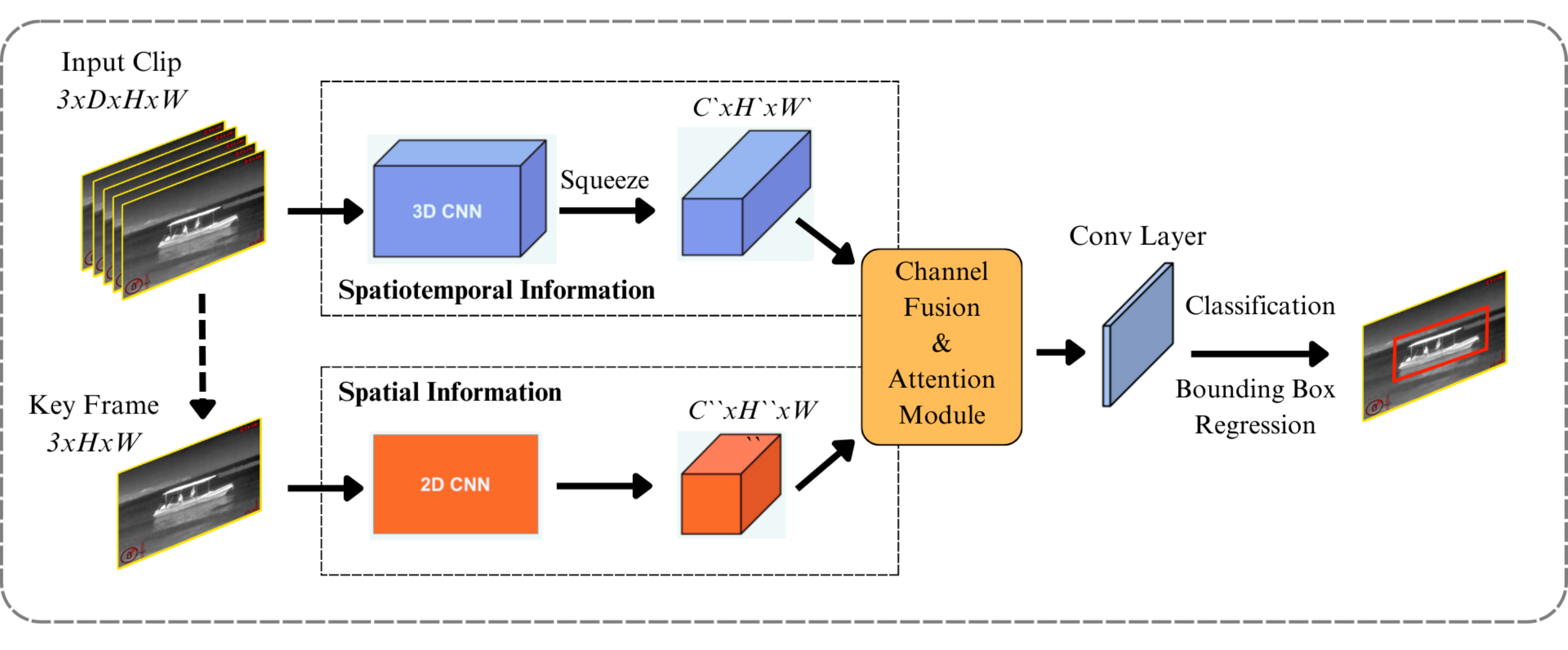}
    \vspace{4pt}
    \caption{Activity detection model architecture}
    \label{activity_picture}
\end{figure*}

Activity detection can be done in both spatial and temporal domains, yet better results yield when both dimensions are considered together. Spatio-temporal action localization is approached in both supervised and semi-supervised techniques. We adapted the YOWO~[\cite{kopuklu2019you}] algorithm which combines the spatial and temporal domain action localization. 
By integrating both information, YOWO effectively captures the dynamics and context of actions in videos. 
It has been originally trained to detect 200 human activities such as walking, talking, running, and cycling in the RGB domain.

As shown in Fig.~\ref{activity_picture}, the algorithm contains two main sections. It captures spatial information and spatiotemporal information separately, and it combines them to do the final classification using channel fusion and an attention module. When reorganizing, we changed the last layer of the model to detect two target activities, swimming and possible human trafficking footage, in our dataset. Then, we retrained the model to detect these two activities in thermal domain using our dataset. To obtain better results, we did a parameter fine tuning by monitoring the frame mAP and video mAP scores. The final values are set as IoU 0.6 and clip length 16. Furthermore, the feeding frame rate for the model is increased from 1 fps to 5 fps.

\subsection{System integration}\label{AA}

In order to facilitate real-time processing, the implementation of two pipelines was deemed necessary as shown in Fig.~\ref{big_picture}. These pipelines operate concurrently, receiving the video feed as input. 
The primary objective of the first pipeline (A and B) is to track and re-identify maritime vessels, while the second pipeline (C) detects suspicious activities within the video stream.
The objects identified by the tracking algorithm are subsequently passed on to the re-identification algorithm to identify detected vessels compared with the database.

The outcome of these pipelines, consisting of processed video feeds, is then channeled into a Graphical User Interface (GUI). This GUI serves as a centralized control interface, enabling users to oversee and manage the entire system. Through the GUI, users can effectively monitor the detected activities and track the identified objects within the video stream in real-time.

\section{Experiments}\label{sec4}

In this section, we describe datasets used, experiment methods and the procedure followed. We used two GeForce RTX 2080 GPUs to infer the system and report the performance indicators mentioned in Table~\ref{Table:Indicators}. We conducted extensive experiments on multiple datasets using several state-of-the-art methods along with our method and the dataset.

\begin{table}[h]
\caption{Key Performance Indicators (KPIs) used to evaluate each algorithm}
\centering
\scriptsize
\begin{tabular}{@{}p{4cm}r@{}}
\toprule
\textbf{Algorithm} & \textbf{Indicator} \\
\midrule
Object detection and tracking & MOT, IDF1, fps \\
Re-identification & Top 1, Top 5, mAP \\
Activity detection & Frame mAP, Video mAP, fps \\
\bottomrule
\end{tabular}
\label{Table:Indicators}
\end{table}

\subsection{Datasets}

\textbf{Our dataset}: Our maritime dataset, captured using a FLIR M232 marine thermal camera, contains videos of moving vessels and maritime objects which are suitable for testing detection and tracking algorithms in the thermal domain. Bounding boxes are drawn for 4 classes including vessels, ships, humans, and jet skies. Furthermore, the dataset contains images of 40 small vessels and 32 large vessels from different viewpoints that can be used to train re-identification algorithms. It contains annotated video feeds of swimming and possible human trafficking activities that can be modeled as suspicious activities.\\
\\
\textbf{VeRi776~[\cite{Veri776}]}: This RGB dataset is a comprehensive collection of vehicle re-identification data, comprising a total of 49,357 images that feature 776 distinct vehicles captured by 20 different cameras. Furthermore, it contains bounding boxes and information regarding vehicle types, colors, and brands. \\
\\
\textbf{VesselID-539~[\cite{qiao2020marine}]}: This RGB dataset is a collection of marine vessel images that were sourced from the website Marine Traffic (www.marinetraffic.com). The raw vessel image dataset encompasses a substantial quantity of data, comprising over 149,465 images representing 539 distinct vessels. On average, each vessel in the dataset is represented by approximately 277 images. \\
\\
\textbf{VehicleID~[\cite{b14}]}: VehicleID dataset contains 26,267 RGB images of vehicles captured from different viewpoints in daytime. For our experiments, we used 500 identities as the training and validation dataset, and another 250 identities as the query and gallery images.\\
\\
\textbf{Singapore Maritime Dataset (SMD)~[\cite{b15}]}: The Singapore Maritime Dataset consists of meticulously curated high-definition near-IR videos captured using strategically positioned Canon 70D cameras around the waters of Singapore. It encompasses on-shore videos acquired from fixed platforms along the shoreline, as well as on-board videos captured from moving vessels, providing diverse perspectives of the maritime environment. This division ensures comprehensive coverage and enables analysis across various viewpoints and scenarios. \\
\\
\textbf{JHMDB-21~[\cite{Jhuang:ICCV:2013}]}: JHMDB is a collection of 960 RGB video sequences featuring 21 different actions for action recognition. It includes video data and annotations for puppet flow, puppet mask, joint positions per frame, action labels per clip, and meta labels per clip.\\

\begin{table*}[t]
    \centering
    \caption{Results of the TraDes algorithm on different domains, evaluated using MOT17, SMD, and Our dataset. Note that, the algorithm has obtained a 61.2\% MOTA score in the IR domain which is the almost same as the RGB domain performance. Therefore, the domain adaptation is successfully achieved while conserving the performance of the algorithm.}
        \begin{tabularx}{\textwidth}{@{}l l l X X X X X X r@{}}
            \toprule
            Dataset & Domain & MOTA$\uparrow$ & IDF1$\uparrow$ & MT$\uparrow$ & ML$\downarrow$ & FP$\downarrow$ & FN$\downarrow$ & IDS$\downarrow$ & FPS$\uparrow$ \\
            \midrule
            MOT17 & RGB & 63.5 & 67.7 & 36.3 & 21.5 & 4.5 & 31.4 & 0.6 & 30 \\
            MOT17 & B\&W & 58.8 & 64.6 & 31.6 & 29.5 & 3.3 & 37.3 & 0.6 & 30 \\
            SMD & Near-IR & 59.5 & 65.2 & 33.1 & 28.1 & 3.4 & 36.6 & 0.6 & 30 \\
            Our & \textbf{IR} & \textbf{61.2} & \textbf{65.4} & 35.5 & 24.9 & 3.6 & 33.2 & 0.6 & \textbf{30}\\
            \bottomrule
        \end{tabularx}
    \label{Table:TraDeS}
\end{table*}

\textbf{UCF101-24~[\cite{b24}]}: UCF101 is a dataset for recognizing actions in real-life RGB videos sourced from YouTube, encompassing 101 action categories. It builds upon the UCF50 dataset~[\cite{ucf50}] and contains 13,320 videos spanning the expanded 101 action categories.\\

Table~\ref{Table:Datasets} summarizes the datasets and methods used for evaluation purposes in the results section.

\begin{table}[h!]
\caption{Experiment catalog}
\centering
\footnotesize
\begin{tabular}{@{}p{2cm}p{1.3cm}r@{}}
\toprule
\textbf{Subsystem} & \textbf{Methods} & \textbf{Datasets used}\\
\midrule
Object tracking & TraDeS & Ours, MOT17, SMD  \\
Re-identification & Our, SPAN & Ours, VeRi776, VehicleID, VesselID-539 \\
Activity detection & YOWO & Ours, JHMDB-21, UCF101-24  \\
\bottomrule
\end{tabular}
\label{Table:Datasets}
\end{table}

\subsection{Training} \label{sec4.2}

{\bf For object detection and tracking},  we trained TraDeS for 4 classes, including vessels, ships, humans, and jet skies. The training was done in two phases. In the first phase, we trained on a subset of classes in the COCO dataset (RGB), relevant to the specific use case, such as vessels and humans. We converted RGB data to grayscale to make COCO images more similar to thermal images. In the second phase, we completely moved to the thermal domain by tuning the model using SMD dataset~[\cite{b15}], along with our thermal data. Subsequently, we fine-tuned the algorithm by adjusting hyperparameters, the learning rate and detection threshold.

{\bf In the re-identification module} as shown in Fig.~\ref{big_picture}(b), we trained area ratio calculation and the feature mapping parts, separately. We trained area ratio calculation using the thermal data of vessels taken from different viewpoints. We used an encoder-decoder model as mentioned in Section~\ref{subsec3.2} for foreground masks extraction.  Then, we trained the model responsible for part attention in SPAN to generate masks for viewpoints using our thermal dataset. In feature mapping part, we use Dino-ViT, which is pre-trained on the ImageNet dataset, for the initial feature extraction. Then, we use transfer learning to train the linear layers in Fig.~\ref{fig_reid_arch} Part (a). In this stage we use our thermal image dataset while keeping the area ratio calculation in the inference as it is already trained.

\section{Results and Discussion}\label{sec5}

In this section, we first evaluate the performance of the TraDes algorithm in thermal domain for maritime object tracking to show that it can obtain similar results as in the RGB domain after the adaptation that we introduced. Second, we show that the view-weighted re-identification approach used in this paper outperforms SPAN method in both RGB and IR domains obtaining higher mAP values. Then, as ablations, we present results with and without view-weighted feature comparison, and effects of CNN and ViT based feature extractions. Furthermore, we evaluate the effect of the number of viewpoints in the feature comparison for the final outcome. Finally, in thermal activity detection, we show that YOWO algorithm performs on par with RGB domain results by evaluating it on JHMDB-21, UCF101-24, and our dataset.


\begin{figure*}[t]
    \centering
    \includegraphics[width=\linewidth]{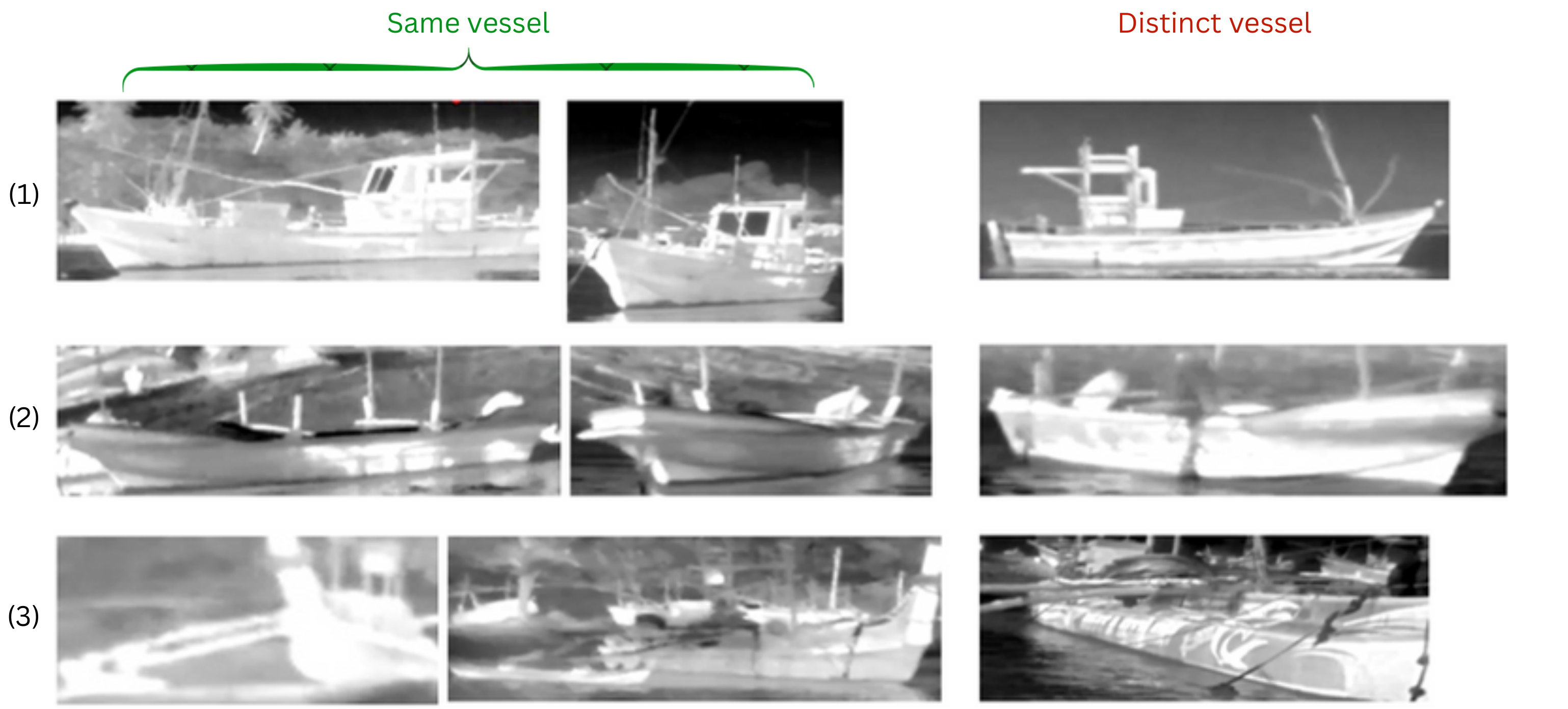}
    \vspace{4pt}
    \caption{Visual results of the re-identification algorithm. 
Each row includes two images depicting the same vessel and one image of a distinct vessel with minor alterations. Our algorithm accurately distinguishes those images in column 2 that pertain to the vessel category in column 1 rather than the vessel category in column 3. Note that the algorithm demonstrates proficiency despite challenges such as orientation variations and blurred images.}
    \label{reid_visual}
\end{figure*}

{\bf Evaluation of our tracking algorithm}:  We evaluated the performance of the TraDes algorithm in the near-IR and IR domains using SMD and our dataset. As shown in Table~\ref{Table:TraDeS}, higher MOTA and MAP scores in our dataset clearly indicate that the algorithm has successfully adapted to the specified classes (vessels, ships, and humans) in the IR domain. The algorithm has obtained a 61.2\% MOTA score in the IR domain which is almost the same as the RGB domain performance. It indicates that we can track objects without color features with only a minor drop in the performance indicators. Also, we could maintain a 15 fps processing speed which is suitable for real-time online tracking. We noticed a considerable drop in the MOTA score for the MOT17 dataset when converted to the black-and-white (B\&W) domain, which can be due to complex and highly dynamic environments with occlusions in the dataset, which is usually not the case for maritime environments. Therefore, the domain adaptation has been successfully achieved while conserving the performance of the algorithm.

{\bf Evaluation of our re-identification algorithm}: 
Our re-identification algorithm convincingly surpasses SPAN and ViT baselines in the IR domain while showing better performance even in the RGB domain with a higher mAP score.
Specifically, in our IR dataset, our algorithm achieved a Top1 accuracy of 81.82\% and a mAP score of 74.26\% compared to SPAN's Top1 accuracy of 78.37\% and mAP score of 73.62\%, indicating the effectiveness of our method in handling infrared images with multiple vessel viewpoints (4.5\% increment in the Top1 score in Table~\ref{Table:reid_results}). Since there are no publicly available thermal vehicle/vessel datasets for re-identification, we conducted experiments with above mentioned RGB datasets to show that our method works competitively in the RGB domain, as well. In the VesselID-539  dataset, our algorithm achieved a Top1 accuracy of 82.60\% (compared to SPAN's 82.43\%), indicating that the proposed method, which is specified for thermal domain performance, is robust in the RGB domain as well. Moreover, across all datasets, our method shows considerably higher mAP scores, outperforming SPAN by 26.7\% on average.

\begin{table*}[h!]
\footnotesize
\centering
\caption{Performance comparison of ReID algorithms. Our algorithm outperforms SPAN in both RGB and thermal domains obtaining the highest mAP socre across all datasets. In particular, in the IR domain that we focus, ours results well-surpass existing results. VW: view weighted.}
\vspace{2mm}
\begin{tabular}{@{}lp{0.9cm}|p{0.8cm}p{0.65cm}p{0.65cm}|p{0.65cm}p{0.65cm}p{0.65cm}|p{0.65cm}p{0.65cm}r@{}}
\toprule
\multicolumn{2}{c}{} &
\multicolumn{3}{>{\centering\arraybackslash}c}{\textbf{SPAN}} &
\multicolumn{3}{>{\centering\arraybackslash}c}{\textbf{ViT Base}} &
\multicolumn{3}{>{\centering\arraybackslash}c}{\textbf{ViT Base + VW (Ours)}} \\
{\bf Dataset} & Domain & {\bf Top1} & {\bf Top5} & {\bf mAP} & {\bf Top1} & {\bf Top5} & {\bf mAP} & {\bf Top1} & {\bf Top5} & {\bf mAP}\\
\midrule
VehicleID & RGB & 78.22 & 85.16 & 56.10 & 71.40 & 82.00  & 71.02 & \textbf{78.40} & \textbf{86.80}  & \textbf{73.62}\\ 

VeRi776 & RGB & \textbf{94.0} & \textbf{97.6} & \textbf{68.9}  & 80.16 & 85.16  & 78.91  & 91.67 & 94.67  & \textbf{80.26} \\ 

VesselID-539 & RGB & 82.43 & 86.67 & 58.10 & 82.52 & 86.70  & 72.18 & \textbf{82.60} & \textbf{86.93}  & \textbf{75.44}\\  
\midrule
Ours & IR & 78.37 & 84.43 & 55.88  & 79.21 & 84.50  & 71.33 & \textbf{81.82} & \textbf{86.36}  & \textbf{74.26}\\
\bottomrule
\end{tabular}
\vspace{-3mm}
\label{Table:reid_results}
\end{table*}

\begin{table*}[h!]
\footnotesize
\centering
\caption{Performance comparison with the number of viewpoints consider in the feature comparison. Including more than one view consistently improves the performance both in RGB and IR domains. The highest improvement is in the IR domain.}
\vspace{2mm}
\begin{tabular}{@{}p{3cm}p{1.5cm}|p{0.9cm}p{0.9cm}p{0.9cm}|p{0.9cm}p{0.9cm}r@{}}
\toprule
\multicolumn{2}{c}{} &
\multicolumn{3}{>{\centering\arraybackslash}c}{\textbf{Largest view}} &
\multicolumn{3}{>{\centering\arraybackslash}c}{\textbf{All views}} \\
{\bf Dataset} & Domain &  {\bf Top1} & {\bf Top5} & {\bf mAP} & {\bf Top1} & {\bf Top5} & {\bf mAP}\\
\midrule
VehicleID & RGB & 71.80 & 82.40  & 71.76 & \textbf{78.40} & \textbf{86.80}  & \textbf{73.62}\\ 

VeRi776 & RGB &  78.83 & 84.83  & 78.56  & \textbf{91.67} & \textbf{94.67}  & \textbf{80.26} \\ 

VesselID-539 & RGB &  72.33 & 78.56  & 71.52 & \textbf{82.60} & \textbf{86.93}  & \textbf{75.44}\\  
\midrule
Ours & IR &  68.18 & 72.72  & 70.75 & \textbf{81.82} & \textbf{86.36}  & \textbf{74.26}\\
\bottomrule
\end{tabular}
\vspace{-3mm}
\label{Table:view_points}
\end{table*}

We explain it using two concepts: (1) The ViT can extract complex features using its attention mechanism which enables paying more attention to specific shapes and masks of vehicles/vessels (Fig.~\ref{reid_visual}). It puts more weight on those features in the feature vector, eventually pushing similar features closer in the feature space. As a result, the mAP score increases as shown in the third main column (ViT Base) of Table~\ref{Table:reid_results}. However, the ViT only cannot maintain a good Top1 score due to the vast variations of the orientation. (2) Secondly, the ArcFace mapping further organizes features in 3 separate spaces (one for each side). It increases the inter-class distances in each side-space and enables side-wise feature comparison, increasing the Top1 and Top5 scores of the algorithm, as shown in the third (ViT Base) and fourth (ViT Base + VW) columns of  Table~\ref{Table:reid_results}. SPAN, in contrast, loses accuracy with increasing orientation changes, resulting in low mAP scores. However, our algorithm is capable of finding matches from the gallery, even with different orientations compared to the query image (Fig.~\ref{reid_visual}), resulting in higher mAP scores. Moreover, we did another ablation study on the number of viewpoints considered in the feature comparison. Here, we considered the side with the highest area ratio as the largest view and did the feature comparison only for that side. As shown in Table~\ref{Table:view_points}, feature comparison in multiple viewpoints (typically 2 views appear in an image) increases the Top1 score by 15\%.

{\bf Evaluation of our activity detection algorithm}: We evaluated the performance of the YOWO algorithm in the B\&W domain by converting UCF101-24 and JHMDB-21 datasets. The algorithm performed well with only minor drops (8.5\%) in performance indicators as shown in Table~\ref{Table:YOWO}. Finally, we evaluated the algorithms for our dataset in the IR domain and obtained a frame mAP score of 72.4\% and a video mAP score of 78.9\%.

\begin{table}[h!]
\footnotesize
\centering
\caption{Performance of the YOWO algorithm in different domains at 5 fps and 0.5 IoU. Note that the domain adapted algorithm has obtained closer results as in the RGB domain after training on limited amount of data.} \vspace{2mm} 
\begin{tabular}{@{}p{1.5cm}p{1cm}p{0.8cm}p{1.2cm}p{1.2cm}r@{}}
\toprule
{\bf Dataset}& {\bf Domain} & {\bf No. of acts} & {\bf Frame mAP} & {\bf Video mAP} & {\bf fps} \\ \midrule
JHMDB-21 & RGB & 21 & 75.7 & 85.9 & 5\\
UCF101-24  & RGB & 101 & 87.3 & 78.6 & 5\\ \midrule
JHMDB-21 & B\&W & 21 & 69.4 & 73.1 & 5\\
UCF101-24  & B\&W & 24 & 76.6 & 71.0 & 5\\
Ours  & \textbf{IR} & 2 & \textbf{72.4} & \textbf{78.9} & \textbf{5}  \\ 
\bottomrule
\end{tabular} \vspace{-3mm}
\label{Table:YOWO}
\end{table}

\section{Conclusion}\label{sec6}

In this paper, we have proposed a thermal vision based approach for maritime surveillance with main contributions in robust vessel re-identification and suspicious activity detection. 
To the best of our knowledge, this is the first time to address maritime vessel re-identification in the thermal domain.  
The adaptation of the TraDeS and the YOWO algorithms for object tracking and activity detection, respectively, was successful in obtaining competitive results as in the RGB domain. In the re-identification algorithm, our novel approach of mapping and comparing features in side-based separate spaces enabled viewpoint independency in re-identification. Our method is proven to be robust for the viewpoint variance in the thermal domain while providing consistent accuracy even with a higher number of classes. It outperformed SPAN algorithm in both thermal and RGB domains. Furthermore, the dataset we created contains images and videos of vessels and suspicious activities that can be used in tracking, activity detection, and vessel re-identification tasks.

Currently, the integrated system works at 2 fps while independent subsystems work at 30 fps for tracking and 5 fps for activity detection. As further developments, a customized hardware setup can be developed for the system for a higher frame rate. Algorithms can be optimized further using parallel computing concepts and obtain a higher throughput. Furthermore, there is a wide research gap in the thermal maritime surveillance domain which should be explored in the future.

\section{Acknowledgment}\label{ack}

Funding for the FLIR thermal camera was provided by the Senate Research Committee Capital Grant: SRC/CAP/2018/02. Computational resources were provided by the Creative Software Pvt. Ltd.

\bibliographystyle{model2-names}
\bibliography{manuscript}

\newpage

\section*{Appendix}
\begin{table*}[t]
\caption{Encoder-decoder architecture for the foreground mask extraction. Here, B refers to the batch size.}
\centering
\begin{tabular}{@{}p{3.5cm}llr@{}}
\toprule
Layer (Type) & Configuration & Input Shape & Output Shape \\ \midrule
InputLayer & - & (B, 192, 192, 3) & (B, 192, 192, 3) \\ \hline
Conv2D & conv2d & (B, 192, 192, 3) & (B, 192, 192, 16) \\ \hline
MaxPooling2D & max\_pooling2d & (B, 192, 192, 16) & (B, 96, 96, 16) \\ \hline
Conv2D & conv2d\_1 & (B, 96, 96, 16) & (B, 96, 96, 8) \\ \hline
MaxPooling2D & max\_pooling2d\_1 & (B, 96, 96, 8) & (B, 48, 48, 8) \\ \hline
Conv2D & conv2d\_2 & (B, 48, 48, 8) & (B, 48, 48, 8) \\ \hline
MaxPooling2D & max\_pooling2d\_2 & (B, 48, 48, 8) & (B, 24, 24, 8) \\ \hline
Conv2D & conv2d\_3 & (B, 24, 24, 8) & (B, 24, 24, 8) \\ \hline
MaxPooling2D & max\_pooling2d\_3 & (B, 24, 24, 8) & (B, 12, 12, 8) \\ \hline
Conv2D & conv2d\_4 & (B, 12, 12, 8) & (B, 12, 12, 8) \\ \hline
UpSampling2D & up\_sampling2d & (B, 12, 12, 8) & (B, 24, 24, 8) \\ \hline
Add & add & [(B, 24, 24, 8), (B, 24, 24, 8)] & (B, 24, 24, 8) \\ \hline
Conv2D & conv2d\_5 & (B, 24, 24, 8) & (B, 24, 24, 8) \\ \hline
UpSampling2D & up\_sampling2d\_1 & (B, 24, 24, 8) & (B, 48, 48, 8) \\ \hline
Add & add\_1 & [(B, 48, 48, 8), (B, 48, 48, 8)] & (B, 48, 48, 8) \\ \hline
Conv2D & conv2d\_6 & (B, 48, 48, 8) & (B, 48, 48, 8) \\ \hline
UpSampling2D & up\_sampling2d\_2 & (B, 48, 48, 8) & (B, 96, 96, 8) \\ \hline
Add & add\_2 & [(B, 96, 96, 8), (B, 96, 96, 8)] & (B, 96, 96, 8) \\ \hline
Conv2D & conv2d\_7 & (B, 96, 96, 8) & (B, 96, 96, 16) \\ \hline
UpSampling2D & up\_sampling2d\_3 & (B, 96, 96, 16) & (B, 192, 192, 16) \\ \hline
Conv2D & conv2d\_8 & (B, 192, 192, 16) & (B, 192, 192, 1) \\ \bottomrule\end{tabular}
\label{tab:enc_dec}
\end{table*}

\clearpage
\end{document}